\documentclass{article}

     \usepackage[nonatbib]{neurips_2020}

\usepackage[utf8]{inputenc} 
\usepackage[T1]{fontenc}    
\usepackage{hyperref}       
\usepackage{url}            
\usepackage{booktabs}       
\usepackage{amsfonts}       
\usepackage{nicefrac}       
\usepackage{microtype}      
\usepackage{graphicx}
\usepackage{array}
\title{Abnormal activity capture from passenger flow of elevator based on unsupervised learning and fine-grained multi-label recognition}

\author{%
  Chunhua Jia \footnotemark[1] \\
  \And
  Wenhai Yi\footnotemark[1] \\
  \And
  Yu Wu\footnotemark[1] \\
  \And
  Hui Huang\footnotemark[1] \\
  \And
  Lei Zhang\footnotemark[1] \\
  \And
  Leilei Wu\footnotemark[1] \\  
}
\begin{document}
\maketitle
\begin{abstract}
  We present a work-flow which aims at capturing residents' abnormal activities through the passenger flow of elevator in multi-storey residence buildings. Camera and sensors (hall sensor, photoelectric sensor, gyro, accelerometer, barometer, and thermometer) with internet connection are mounted in elevator to collect image and data. Computer vision algorithms such as instance segmentation, multi-label recognition, embedding and clustering are applied to generalize passenger flow of elevator, i.e. how many people and what kinds of people get in and out of the elevator on each floor. More specifically in our implementation we propose GraftNet, a solution for fine-grained multi-label recognition task, to recognize human attributes, e.g. gender, age, appearance, and occupation. Then anomaly detection of unsupervised learning is hierarchically applied on the passenger flow data to capture abnormal or even illegal activities of the residents which probably bring safety hazard, e.g. drug dealing, pyramid sale gathering, prostitution, and over crowded residence. Experiment shows effects are there, and the captured records will be directly reported to our customer(property managers) for further confirmation.
\end{abstract}

\section{Introduction}
In modern city, most people live in condos or apartments of multi-storey buildings. Considering complexity of structure and high density of residents, public safety is challenged in such buildings[1,2]. Technologies based on Artificial Intelligence(AI) and Big Data and Internet Of Things(IoT) make it possible to capture or predict some behaviors/activities/events[3,4,5] which have direct or potential safety hazard to the residents, and thus precautions can be taken accordingly. Meanwhile, people's privacy is another big consideration[6]. Even it is for public safety, still we need to make sure that all data is properly collected and used, and thus privacy of the residents will not be violated. On the other hand, since patterns of behaviors or activities conducted by highly socialized citizens vary from one to one and change continuously, it's particularly difficult to give a specific definition on residents' activities which would put their own safety in danger. Considering all the aspects above, we find that elevator could be the most feasible and suitable environment to take operation on because it's legal and reasonable to deploy public surveillance and people take elevator widely and frequently enough.

Anomaly detection[7], which is also known as outlier or novelty detection, is a widely studied topic that has been applied to many fields including medical diagnosis, marketing, network intrusion, and many other applications except for automated surveillance. To capture activities with potential public safety hazard, anomaly detection of unsupervised learning is our choice from algorithm perspective because the definition of safety hazard on people's activity is ambiguous and we assume such activities are relatively rare.

Similar to [8,9,10,11,12], the initial goal of our system is to capture elevator malfunction (e.g. stuck) in real-time through setting several different sensors in elevator and analyzing data collected. A camera is mounted and used to automatically check if any passenger is trapped in elevator box when malfunction happens through Computer Vision(CV) algorithm, e.g. pedestrian detection. Elevator malfunction with passenger trapped will be directly reported to the manager of the residence building with the highest priority so that rescue will be carried out as soon as possible. The system introduced above is deployed on more than 100000 elevators currently. Based on this system, we develop a framework to capture abnormal activities of residents.

\section{Related Works}
Some researchers focus on predictive maintenance based on detection of abnormal usage of an elevator, taking advantage of sensors information[13]. However, common Internet Of Things(IoT) solution can only detect hardware malfunction, but is unable to describe the behavior of residents. Video surveillance fits this task better. There are many surveys conducted to anomalous events detection which utilize video surveillance. Hu[14] gave a survey on visual surveillance of object motion and behaviors, and proposed a general processing framework including several stages such as environment modeling, motion segmentation, object classification, understanding and description of behaviors, which all belong to computer vision domain. Surveillance system should become more intelligent, crucial, and comprehensive to deal with the situations under which individual safety could be compromised by potential criminal activity. Tomi[15] analyzed three generations of contemporary surveillance system and the most recent generation is decomposed into multi-sensor environments, video and audio surveillance, and distributed intelligence and awareness.

Many surveys of anomaly detection used supervised learning algorithm[16,17,18,19,20], which are based on the assumption that normal and abnormal behavior can be well distinguished. However, anomalous behavior could not be clearly defined, resulting in insufficient labeled data for supervised model.  Xiang[21,22] developed a runtime accumulative anomaly measurement for behavior captured in videos, which is based on an online Likelihood Ratio Test(LRT) method and shows better performance on unlabeled data set.

Research[23] defines three common assumptions of anomalous behavior: anomalous events occur infrequently comparing to normal events, have significantly different characteristics from normal events, and have a specific meaning. These ideas inspire our work. We utilize these ideas to capture abnormal activity of residents in the building. For example, if the passenger flow of one floor is much higher than others, this floor may have anomalous events or activities. Our system tends to capture abnormal pattern of passenger flow using anomaly detection algorithm.

Anomaly detection is the identification of data sections which significantly differ from a regular or normal pattern. There are three kinds of anomaly detection algorithms. The first kind is based on statistic method that usually build a probability distribution model and then calculate corresponding probability to choose object with low probability as anomalous data points. The second is clustering method based on the distribution density of data features. Clusters with different distribution characteristics from other are likely to be considered as anomalous data cluster[24,25]. The third is specialized anomaly detection method represented by One Class SVM and Isolation Forest[26,27], which does well in abnormal points detection. One Class SVM is a novelty detection method other than outlier detection which usually utilizes normal data points for training and then deploys trained model to find abnormal data points. Isolation Forest was proposed by students of Zhi-Hua Zhou. It mainly uses the idea of integrated learning to detect abnormal points, and has almost become the first choice of abnormal point detection algorithm.

\section{Passenger flow collection through AI and IoT}
\subsection{Infrastructure of internet of things (IoT) for data collection}
Camera and several kinds of sensors (Hall sensor, photoelectric sensor, gyro, accelerometer, barometer, thermometer, etc.) are mounted in elevator as the deployment of our system. As a solution which is easy to promote with low cost, our system does not require any access to the elevator's ECU or other electronic signals, which means our solution could be deployed in most of the existing elevators. With the sensors properly setup, we are able to monitor the trace of elevator running. To collect passenger data, we use images captured from the camera instead of video stream. We take one snapshot each time during the elevator moves from one floor to another. Alongside with the image, time stamp and floor numbers of start/end are also collected. Images with extra information are uploaded and stored on cloud for further analysis. Such mechanism of snapshooting is based on one simple assumption: people get in or out of the elevator only when it stops at certain floor, and thus we can generalize how many and what kinds of people get in and out with comparing images successively captured, as shown in Figure 1.
\begin{figure}[ht]
	\centering
	\includegraphics[scale=0.3]{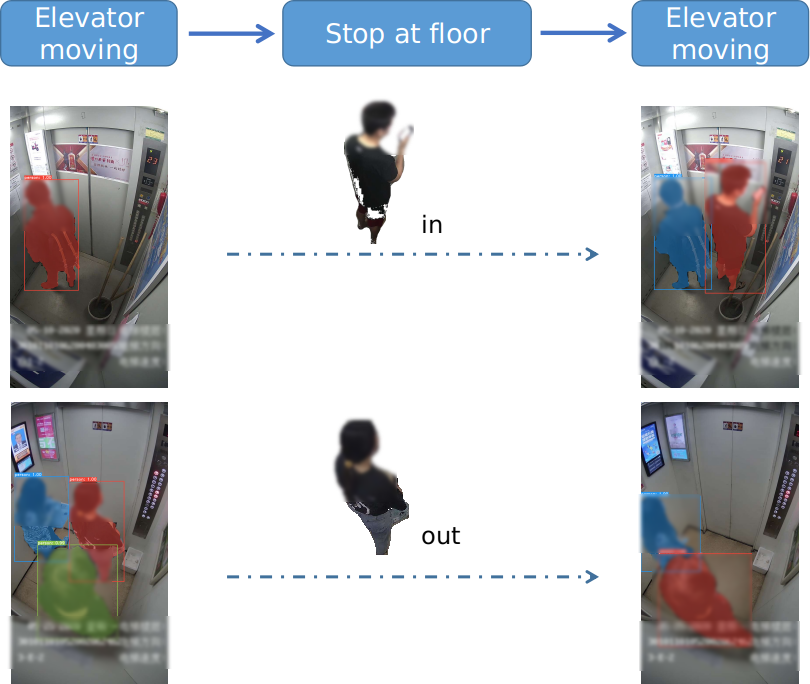} 
	\caption{Comparison of two images successively taken.}
\end{figure}

\subsection{Instance segmentation, embedding and clustering}
Considering occlusion is inevitable when there are several passengers in the elevator at the same time, for feature extraction on each individual passenger, instance segmentation is needed. Different from object detection, instance segmentation can accurately segment all objects at pixel level and minimize the impact of occlusion and background. It could be considered as a pre-process similar to attention mechanism, so that other CV models could focus on human target itself completely.

YOLACT[28], a representative one-stage method which was proposed to speed up instance segmentation, is utilized by us to segment target person from background and other non-targets as shown in Figure 2. 

\begin{figure}[ht]
	\centering
	\includegraphics[scale=0.3]{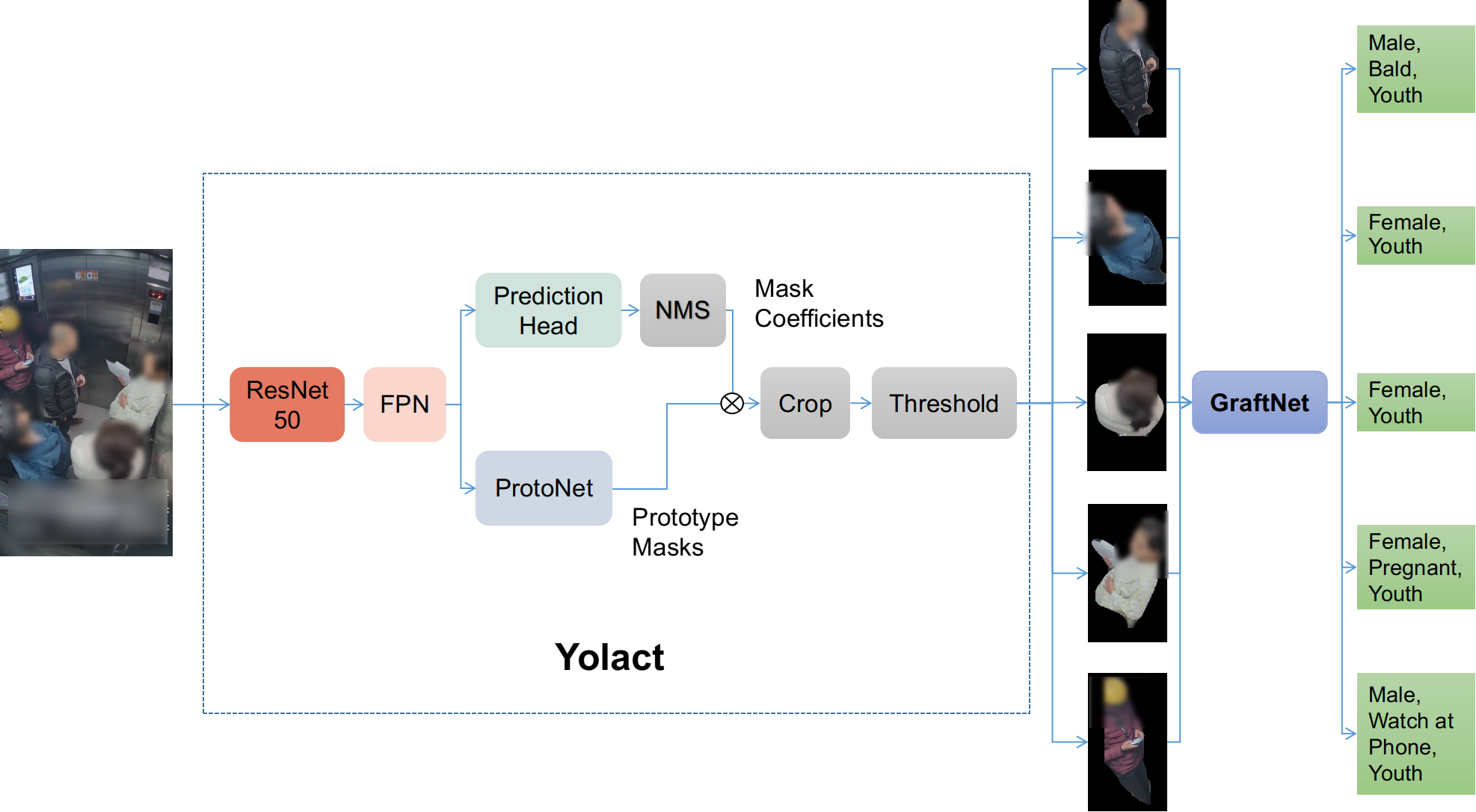} 
	\caption{ Architecture of YOLACT.}
\end{figure}

As mentioned above, the information of people getting in and out of each floor could be generalized by comparing two successively captured images. For that, passengers in such pair of images need be vectorized and association between them has to be established.

We assume the variance of passengers' overall appearances is greater than that of their faces, so FaceNet[29] designed for face embedding and clustering should be sufficient to identically represent the appearance of each segmented individual. As a result, FaceNet is re-trained with our own dataset of segmented passengers and utilized as feature extractor.

As shown in Figure 3, there are \emph{M} passengers in image \emph{p1} captured right before the elevator stops at certain floor, and \emph{N} passengers in image \emph{p2} captured right after the elevator leaves that floor. All those segmented passengers are fed to FaceNet and corresponding feature vectors are returned. Then we build an association matrix \emph{D} with order \emph{M * N}. The element \emph{d$_{ij}$} represents the Euclidean distance between the feature vectors of passenger \emph{i} in \emph{p1} and passenger \emph{j} in \emph{p2}. Minimal value searching on each row (or column) with a threshold \emph{t} is carried out on \emph{D} to get the best match for each passenger (no match found if the minimal value is greater than \emph{t}). The passengers in \emph{p2} without any match from \emph{p1} are considered as "get out of that floor" and those in \emph{p1} without match from \emph{p2} are considered as "get into the floor". Then attributes recognition are only applied on above two kinds of passengers for each floor.

\begin{figure}[ht]
  \centering
  \includegraphics[scale=0.3]{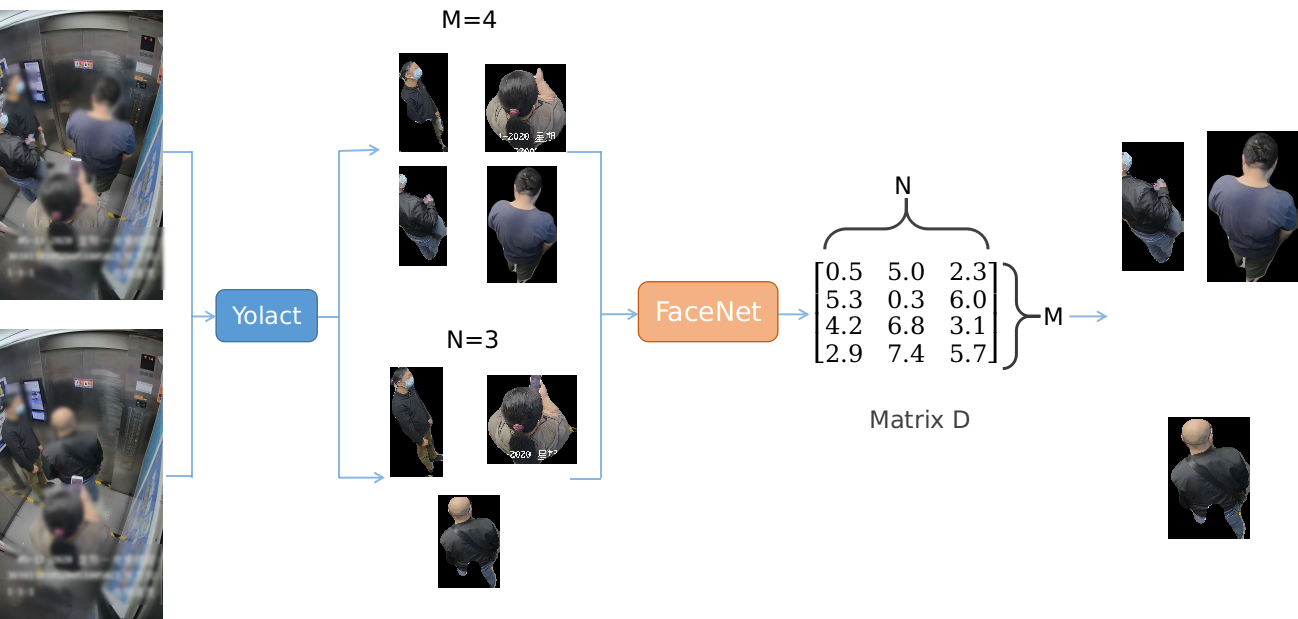} 
  \caption{Segmentation, embedding and clustering to generalize passenger flow of each floor.}
\end{figure}

\subsection{GraftNet: a solution for fine-grained multi-label classification}
To analyse the passenger flow of elevator in residence building, data with more descriptive information is needed, e.g. gender, age, occupation, appearance, etc. Here we propose GraftNet---a solution for fine-grained multi-label recognition task, i.e. to recognize different attributes of elevator passengers in our case.

\begin{figure}[ht]
  \centering
  \includegraphics[scale=0.4]{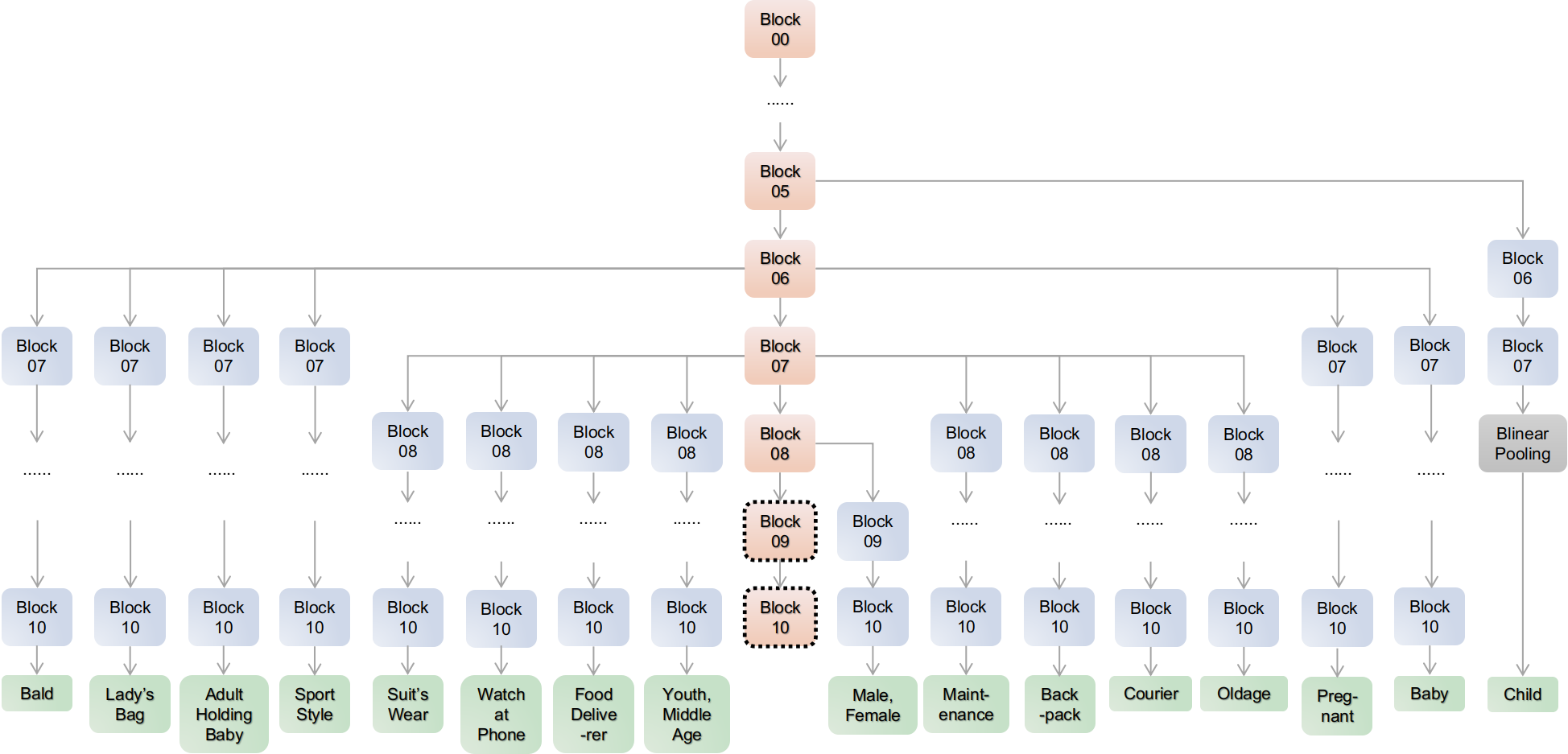} 
  \caption{Architecture of GraftNet (based on Inception V3).}
\end{figure}

GraftNet is a tree-like network that consists of one trunk and branches corresponding to attributes, as shown in Figure 4. The trunk is used to extract low-level features such as shapes and textures, which can be commonly represented with generic features. And branches is mainly used to generate high-level features and thus can be customized for different attributes. For GraftNet, we propose a two-step training procedure. Instead of to annotate all samples for all attributes overall, samples are collected and annotated for one single attribute separately, so that we get the sub-datasets corresponding to attributes. In first step, InceptionV3 is pre-trained on the collection of all sub-datasets by using a dynamic data flow graph, as shown in Figure 5. The 11 blocks with pre-trained weights could be considered as the trunk of GraftNet. The second step is to separately fine-tune and graft branches onto the trunk for each attribute. By training trunk and branches in a two-step way, GraftNet could save time and labor for both annotation and training. Sub-datesets of different attributes could be maintained separately and incrementally, i.e. new attributes or samples could be added without any re-work on the existing data set. Besides, training task of one-branch-for-one-attribute (the iteration of samples re-collecting and model fine-tuning) is more manageable in practice for a team-work. So to speak, the very basic consideration of the design of GraftNet is that the requirement of recognizing a new attribute could come at any time and we don't want any re-work because of that.

\begin{figure}[ht]
  \centering
  \includegraphics[scale=0.6]{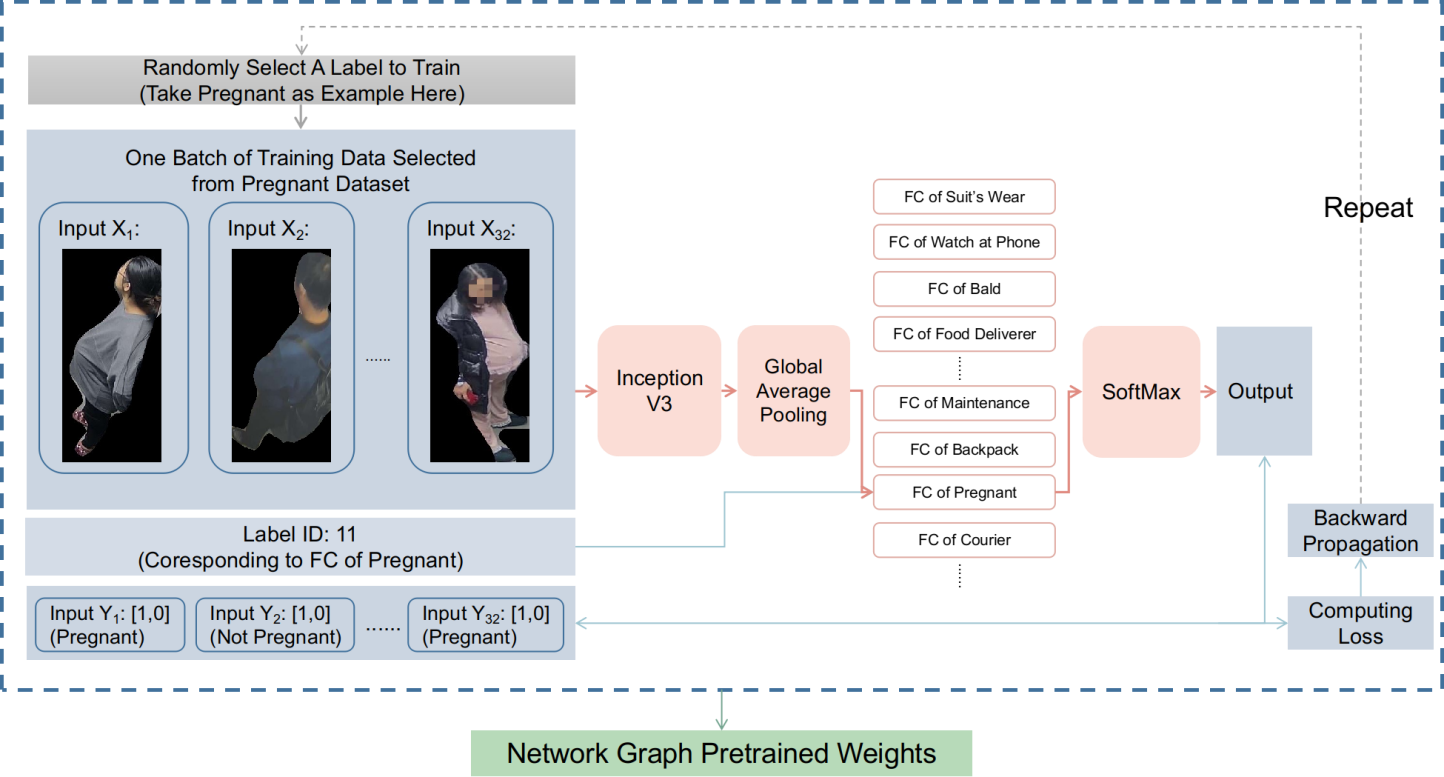} 
  \caption{GraftNet: Process of training with dynamic data flow graph.}
\end{figure}

Experimental results show that GraftNet performs well on our human attributes recognition task (fine-grained multi-label classification), and the combination of pre-trained trunk and fine-tuned branches can effectively improve the accuracy, as shown in Figure 6 and Figure 7.

\begin{figure}[ht]
	\centering
	\includegraphics[scale=0.3]{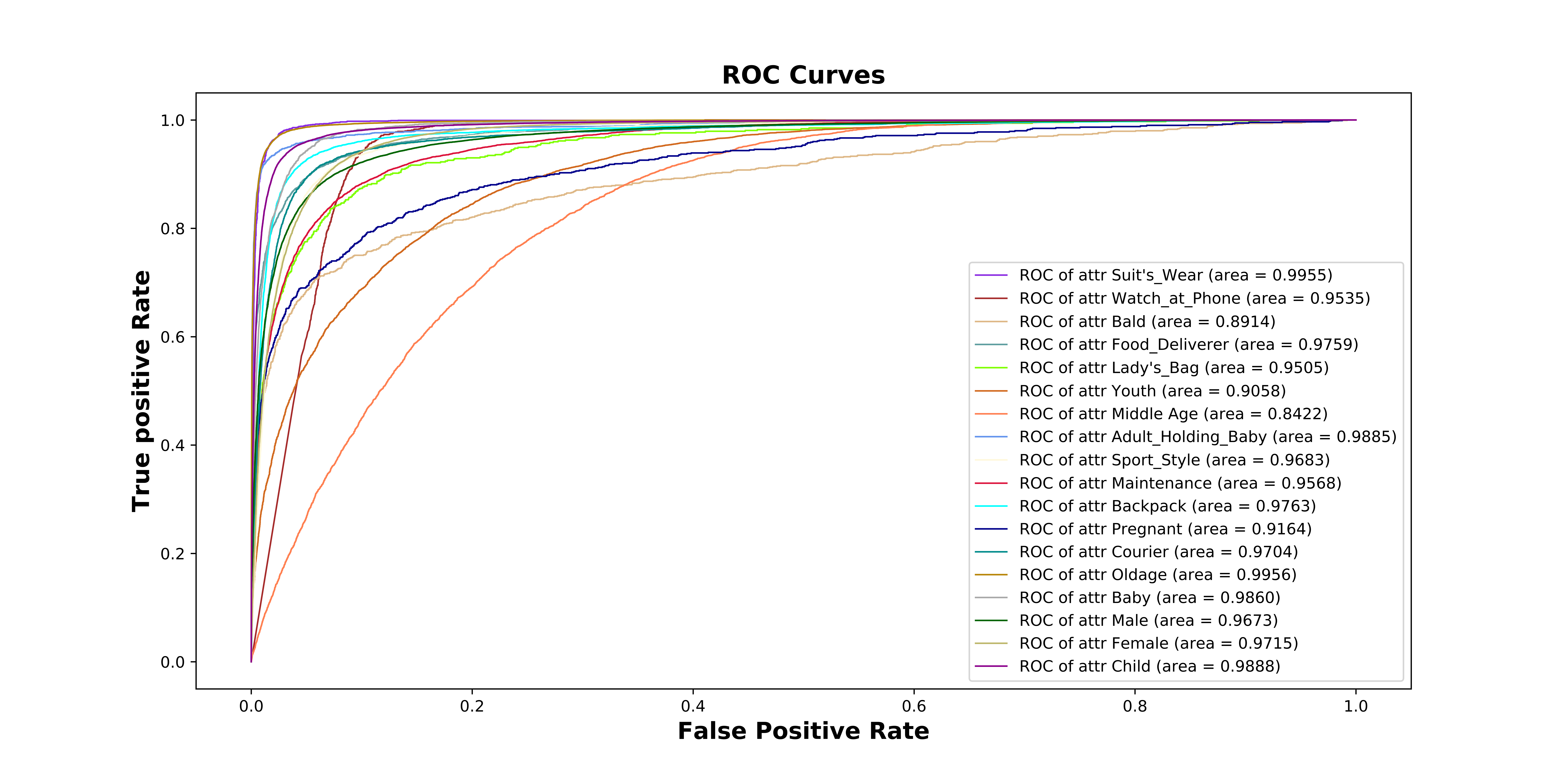} 
	\caption{ROC curves of attributes with our pretrained weights.}
\end{figure}

\begin{figure}[ht]
	\centering
	\includegraphics[scale=0.3]{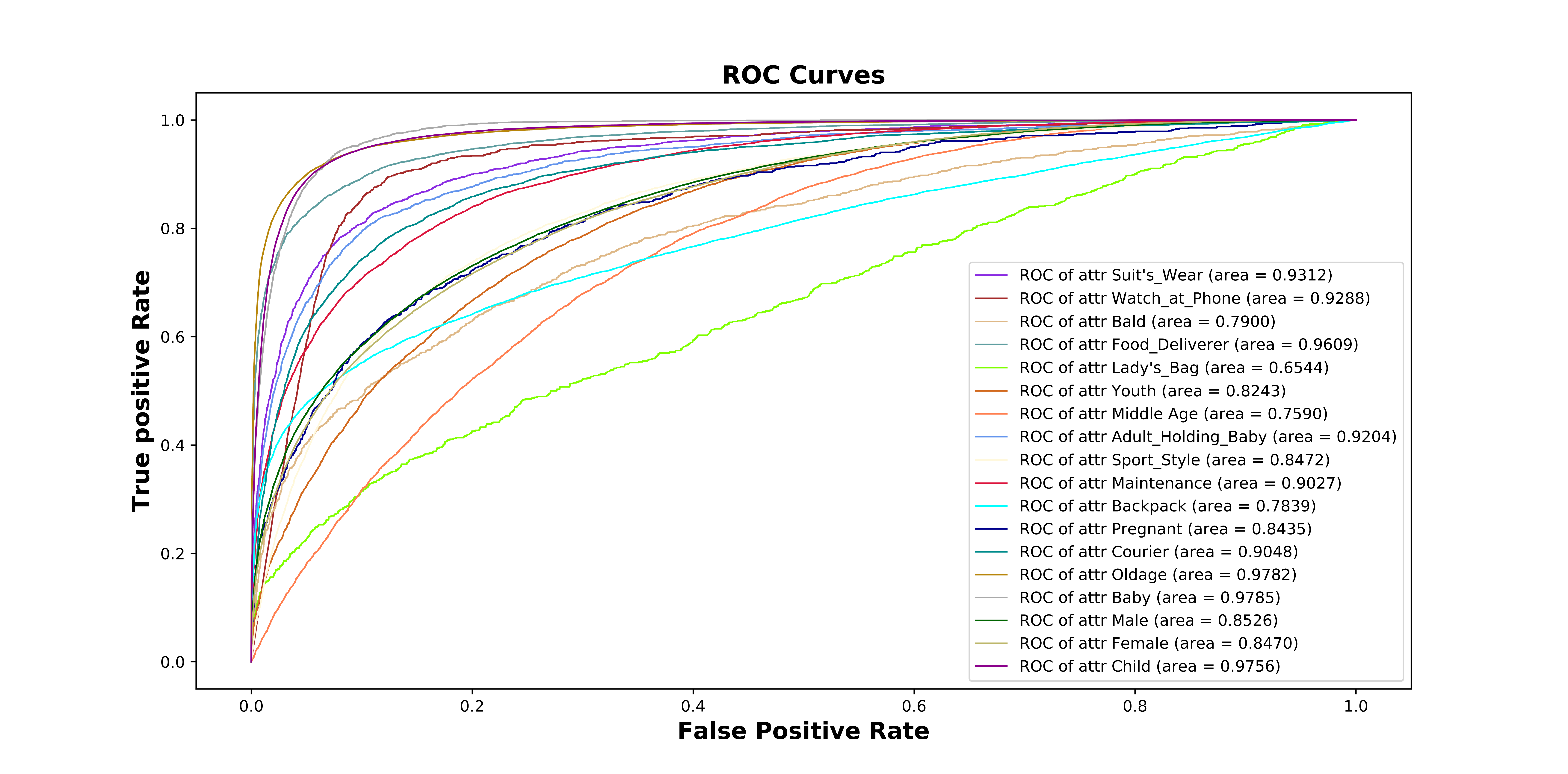} 
	\caption{ROC curves of attributes with pretrained weights on ImageNet.}
\end{figure}

Since GraftNet is deployed on cloud but not embedded devices in our system, we focus more on extendability rather than efficiency. From some perspectives, our work is quite like an inverse process compared to network pruning or weights compression[30,31]. Rather than to reduce the redundancy of neural networks for a fixed task, what we do is to leverage the over-parameterization and maximum its usage to recognize new attributes with a few extra branches.

\section{Hierarchical anomaly detection for abnormal activity capture}
\subsection{Isolation Forest: anomaly detection of unsupervised learning}

Isolation forest(IForest) is an unsupervised learning algorithm dealing with anomaly detection problem. Instead of building a model of normal instance like common techniques use, the principle of Isolation Forest is isolating anomalous points which means the tendency of anomalous instances can be easier to be split from the rest samples in a data set, compared with normal points, because anomalous data points has two quantitative properties: different and fewer than normal data points.

Anomaly detection with Isolation Forest is a process composed of two main stages: In the first stage, a training data set is used to build isolation trees(iTrees). In the second stage, each instance in test set is passed through the iTrees build in the previous stage, and a proper “anomaly score” is assigned to the instance according to the average path length of iTrees. Once all the instances in the test set have been generated an anomaly score, it is possible to mark the point whose score is greater than a predefined threshold as anomaly.

As one of the most famous anomaly detection algorithms, IForest has outstanding advantages: It has a low linear time complexity and a small memory requirement. IForest can be used in huge amounts of data sets because of its basic approach of random forest. The independence of Isolation trees ensures this model can be employed on large scale distributed system to accelerate Computing Platform. At the same time, large number of Isolation trees makes algorithm more stable. 

\subsection{Feature generation and hierarchical anomaly detection}
For the generalized passenger flow we mentioned in Section 3, we treat them as two parts hierarchically. The first is the result of instance segmentation, embedding and clustering, which is called flow-count data here, i.e. how many people get in/out of elevator. And the other is the result of human attributes recognition, which is called attributes data, i.e. what kinds of people get in/out of elevator. The hierarchy is defined here because the flow-count data is actually byproduct of our system in commercial use but attributes data is not. Since our IoT system is deployed in more than 100000 elevators which transport tens of millions people daily, to additionally recognize attributes for all these passengers is computationally not affordable for us. Besides, the original requirement from our customer at the very beginning is to find out over crowded residence (e.g. more than 20 illegal migrants live in one small apartment), it's reasonable that we put more consideration on flow-count data. Therefore we decide to perform anomaly detection hierarchically, first on flow-count data then on attributes data.

Flow-count data is used for the first round of anomaly detection. For the weekdays in the last 15 days, we calculate the mean \emph{m1} and standard deviation \emph{s1} of passenger flow per floor per elevator, \emph{m2} and \emph{s2} of the flow per elevator, and \emph{m3} and \emph{s3} of the flow of all elevators in the same residential estate. Considering citizens are highly civilized and socialized, \emph{m4$\sim$m6} and \emph{s4$\sim$s6} are calculated for the weekends in the last 15 days, the same as \emph{m1$\sim$m3} and \emph{s1$\sim$s3} on weekdays. Including the floor number, we get feature vectors of length \emph{13} per floor per elevator.

The contamination parameter of isolation forest during training procedure prescribes the proportion of outliers in the data set. We set it as 0.2 to output as many records as possible for the next second round anomaly detection.

Attributes recognition is only performed on the output of the first round anomaly detection to reduce computation. Besides the 12 values of mean and standard deviation generated in first round (floor number excluded), the attribute features are adopted by calculating the mean of attribute recognition result per floor per elevator. In detail, for passengers who get in/out elevator of certain floor, attributes recognition with GraftNet is performed to get feature vectors (22 classification results and 22 corresponding scores), then mean of these attribute feature vectors are calculated. The head count and distribution of time of appearance in 24-hour are also included. As a result, for the second round anomaly detection, we get feature data of length 81 as shown in Table 1.

\begin{table}[ht]
\centering
\caption{Feature data for anomaly detection}
\centering
\begin{tabular}{p{0.3cm}p{0.2cm}p{0.3cm}|p{0.5cm}p{0.2cm}p{0.8cm}|c|cp{0.2cm}p{0.6cm}|p{0.5cm}p{0.2cm}p{0.6cm}|p{0.4cm}p{0.2cm}p{0.7cm}}
\hline
\multicolumn{6}{c|}{flow count}&
\multicolumn{10}{c}{attributes}\\
\hline
\multicolumn{3}{c|}{mean}&
\multicolumn{3}{c|}{std}&count&
\multicolumn{3}{c|}{classification}&
\multicolumn{3}{c|}{score}&
\multicolumn{3}{c}{24-hour}\\
\hline
m1&...&m6&s1&...&s6&Num&t1&...&t22&ts1&...&ts22&h1&...&h24\\
\hline
141&...&131&22.31&...&176.86&666&0.004&...&0.01&0.015&...&0.018&0&...&0.009\\
140&...&26&92.86&...&20.46&201&0.005&...&0.06&0.014&...&0.015&0&...&0.005\\
133&...&93&44.78&...&64.65&607&0.003&...&0.005&0.017&...&0.008&0.024&...&0.024\\
129&...&131&21.80&...&176.86&555&0&...&0.02&0.006&...&0.013&0.014&...&0.019\\
\hline
\end{tabular}
\end{table}

The contamination parameter of isolation forest is set as 0.01 for the second round detection. There are around 1 million records for each floor from 100000 elevators. In our experiment, after two rounds of anomaly detection, we finally get 643 outliers output.

\subsection{Manual review and analysis}
One fact that we have to admit is that sometimes the outliers obtained from anomaly detection of unsupervised learning might not be activities with public safety hazard. The anomaly of such outliers could be caused by malfunction of IoT system (e.g. malfunction of sensors, camera, network), or misentries in our database. For example, most of the elevators in our system are from residence buildings, and very few are from non-residence buildings, such as office building, hospital, school, and shopping mall. The activities of people in such non-residence buildings are obviously different from residence buildings. Our abnormal activity capture is supposed to perform on residence buildings only. However, some non-residence buildings are misentried as residence ones in our system, which could probably make them captured as anomaly. Similarly malfunction of IoT system could also cause exception data which might be captured.

\begin{figure}[ht]
	\centering
	\includegraphics[width=13.5cm]{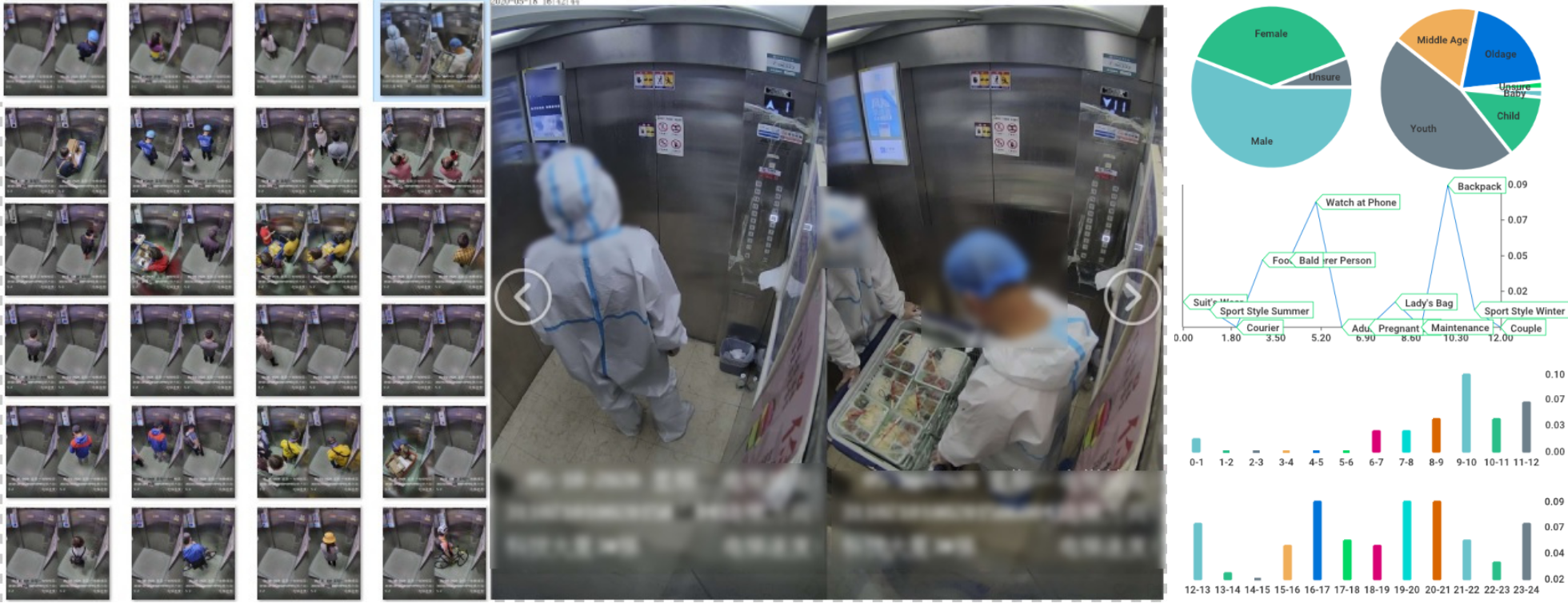} 
	\caption{Interface of inspection tool.}
\end{figure}

To verify the outliers and keep improving the result, we build inspection tools to review and analyze the output manually. Corresponding to each output, the image-pairs successively captured will be reviewed, together with the statistical data such as distribution of attributes and time of appearance in 24 hours, as shown in Figure 8. All data exceptions caused by malfunction/misentries and abnormal activities clearly without any public safety hazard (e.g. running company in home office, decorators getting in/out, etc.) will be logged and excluded from next round of anomaly detection. Meantime confirmed records with suspicions of safety hazard will be reported to our customer.

\section{Experiments and conclusion}
Out of the 643 records output in our experiment we randomly pick 412 and review them one by one with the inspection tool. The result is shown in Table 2.

\begin{table}[ht]
\centering
\caption{Results after reviewed}
\begin{tabular}{ccp{5.5cm}c}
\hline
review\_result&safety hazard&\multicolumn{1}{c}{comments}&number\\
\hline
positive&probably yes/unsure&\multicolumn{1}{m{7cm}}{Something different is there. Need to be checked by property manager.}&289\\
\hline
positive&no&caused by malfunction of sensors&13\\
\hline
positive&no&apartment under decoration&32\\
\hline
positive&no&dormitory/hotel&27\\
\hline
positive&no&shopping mall/entertainment venue&2\\
\hline
positive&no&office building&40\\
\hline
positive&yes&catering service running in apartment&3\\
\hline
positive&yes&over crowded residence&6\\
\hline
\end{tabular}
\end{table}

As for now we don't have any way to directly validate this framework to get a definite conclusion because it hasn't been put into commercial use, i.e. it's still experimental and no records are reported to the property manager and verified by them. And once this framework is put into use in the future, more and more exceptions (anomalous but no safety hazard) will be captured and excluded from the periodic running, and finally actual safety hazard will be captured more accurately.

\section*{References}
\small

[1] Hanapi, N. L. , Ahmad, S. S. , Ibrahim, N. , Razak, A. A. , \& Ali, N. M. (2017). Suitability of escape route design for elderly residents of public multi-storey residential building. Pertanika Journal of Social Science and Humanities, 25(s)(2017), 251-258.

[2] Hanapi, N. , Ahmad, S. , \& Abd Razak, Azli. (2019). EMERGENCY SAFETY FOR MULTI-STOREY PUBLIC HOUSING IN KUALA LUMPUR. 8. 64-70.

[3] Almeida, Aitor, Azkune, \& Gorka. (2018). Predicting human behaviour with recurrent neural networks. Applied Sciences.

[4] Hartford J. , Wright James R. , Leyton-Brown K. (2016). Deep Learning for Predicting Human Strategic Behavior. in proceedings of NIPS 2016.

[5] Luceri, L. , Braun, T. , \& Giordano, S. (2019). Analyzing and inferring human real-life behavior through online social networks with social influence deep learning.

[6] Jain, P. , Gyanchandani, M. , \& Khare, N. (2016). Big data privacy: a technological perspective and review. Journal of Big Data. 3. 10.1186/s40537-016-0059-y. 

[7] Chandola, V. , Banerjee, A. , \& Kumar, V. (2009). Anomaly Detection: A Survey. ACM Comput. Surv.. 41. 10.1145/1541880.1541882. 

[8] Huang, M. , Liu, Z. , \& Tao, Yang. (2019). Mechanical Fault Diagnosis and Prediction in IoT Based on Multi-source Sensing Data Fusion. Simulation Modelling Practice and Theory. 101981. 10.1016/j.simpat.2019.101981. 

[9] Zihan, M. , Shaoyi, H. , Zhanbin, Z. , \& Shuang, X. . (2018). Elevator safety monitoring system based on internet of things. International Journal of Online Engineering, 14(08).

[10] Lai, C. T. , Jackson, P. R. , \& Jiang, W. (2017). Shifting paradigm to service-dominant logic via internet-of-things with applications in the elevators industry. Journal of Management Analytics, 4(1), 35-54.

[11] Wan, Z. , Yi, S. , Li, K. , Tao, R. , Gou, M. , \& Li, X. , et al. (2015). Diagnosis of elevator faults with ls-svm based on optimization by k-cv. Journal of electrical and computer engineering, 2015, 935038.1-935038.8.

[12] Li, Y. , \& Zhang, H. H. (2010). Intelligent elevator detecting system based on neural network. journal of beijing university of technology.

[13] You Z. , Kai W. , \& Hongxia L. (2018). An elevator monitoring system based on the internet of things. procedia computer science, 131, 541-544.

[14] Hu, W. , Tan, T. , Wang, L. , \& Maybank, S. (2004). A survey on visual surveillance of object motion and behaviors. IEEE Transactions on Systems Man and Cybernetics Part C (Applications and Reviews), 34(3), 334-352.

[15] Tomi D. Räty. (2010). Survey on contemporary remote surveillance systems for public safety. IEEE Transactions on Systems, Man, and Cybernetics, Part C (Applications and Reviews), 40(5), 493-515.

[16] Dee, H. , \& Hogg, D. . (2004). Detecting inexplicable behaviour. 10.5244/C.18.50. 

[17] Duong, T. , Bui, H.H. , Phung, Dinh \& Venkatesh, Sriram. (2005). Activity Recognition and Abnormality Detection with the Switching Hidden Semi-Markov Model. Proceedings - 2005 IEEE Computer Society Conference on Computer Vision and Pattern Recognition, CVPR 2005. 1. 838- 845 vol. 1. 10.1109/CVPR.2005.61. 

[18] Gong, S. , \& Xiang, T. (2003). Recognition of group activities using dynamic probabilistic networks. IEEE International Conference on Computer Vision. IEEE.

[19] Morris, R.J. \& Hogg, David. (1998). Statistical Models of Object Interaction. International Journal of Computer Vision. 37. 81-85. 10.1109/WVS.1998.646024. 

[20] Oliver, N. M. , Rosario, B. , \& Pentland, A. P. (2000). A bayesian computer vision system for modeling human interactions. IEEE Trans. Pattern Anal. Mach. Intell, 22(8), 831-843.

[21] Xiang, T. , \& Gong, S. (2008). Video behavior profiling for anomaly detection. IEEE Transactions on Pattern Analysis and Machine Intelligence, 30(5), p.893-908.

[22] Xiang, T. , \& Gong, S. (2005). Video behaviour profiling and abnormality detection without manual labelling. Proceedings of the IEEE International Conference on Computer Vision. 2. 1238- 1245 Vol. 2. 10.1109/ICCV.2005.248. 

[23] Sodemann, A. A. , Ross, M. P. , \& Borghetti, B. J. (2012). A review of anomaly detection in automated surveillance. IEEE Transactions on Systems Man \& Cybernetics Part C, 42(6), 1257-1272.

[24] Angiulli, F. , Basta, S. , \& Pizzuti, C. (2006). Distance-based detection and prediction of outliers. IEEE Transactions on Knowledge and Data Engineering, 18(2), 145-160.

[25] Jin, W. , Tung, A. , Han, Jiawei \& S, Canada. (2001). Mining Top-n Local Outliers in Large Databases. Proceedings of the Seventh ACM SIGKDD International Conference on Knowledge Discovery and Data Mining. 10.1145/502512.502554. 

[26] Liu, F. T. , Ting, K. M. , \& Zhou, Z. H. (2012). Isolation-based anomaly detection. Acm Transactions on Knowledge Discovery from Data, 6(1), 1-39.

[27] Liu, F. T. , Ting, K. M. , \& Zhou, Z. H. (2009). Isolation Forest. Data Mining, 2008. ICDM '08. Eighth IEEE International Conference on. IEEE.

[28] Bolya, D. , Zhou, C. , Xiao, F. , \& Lee, Y. J. (2019). Yolact: real-time instance segmentation.

[29] Schroff, F. , Kalenichenko, D. , \& Philbin, J. (2015). FaceNet: A unified embedding for face recognition and clustering. 2015 IEEE Conference on Computer Vision and Pattern Recognition (CVPR). IEEE.

[30] Ullrich, K. , Meeds, E. , \& Welling, M. (2017). Soft weight-sharing for neural network compression.

[31] Han, S. , Mao, H. , \& Dally, W. J. (2016). Deep Compression: Compressing Deep Neural Networks with Pruning, Trained Quantization and Huffman Coding. ICLR.

\end{document}